\documentclass[10pt,twocolumn,letterpaper]{article}

\usepackage{iccv}              

\definecolor{iccvblue}{rgb}{0.21,0.49,0.74}
\usepackage[pagebackref,breaklinks,colorlinks,allcolors=iccvblue]{hyperref}

\def\paperID{8876} 
\def\confName{ICCV}
\def\confYear{2025}

\title{{MamTiff-CAD: Multi-Scale Latent Diffusion with Mamba+ for Complex Parametric Sequence}}

\author{
Liyuan Deng\textsuperscript{1} \quad 
Yunpeng Bai\textsuperscript{2} \quad 
Yongkang Dai\textsuperscript{1} \quad 
Xiaoshui Huang\textsuperscript{3} \quad 
Hongping Gan\textsuperscript{1} \\
Dongshuo Huang\textsuperscript{1} \quad 
Hao Jiacheng\textsuperscript{4} \quad 
Yilei Shi\textsuperscript{1}\thanks{Corresponding author.} \\
\textsuperscript{1}Northwestern Polytechnical University \quad 
\textsuperscript{2}National University of Singapore \\
\textsuperscript{3}Shanghai Jiao Tong University \quad 
\textsuperscript{4}Nanchang University \\
{\tt\small \{dly,daiyongkang,ganhongping,huangdongshuo,yilei\_shi\}@mail.nwpu.edu.cn} \\
{\tt\small bai\_yunpeng99@u.nus.edu \quad huangxiaoshui@sjtu.edu.cn \quad haojiacheng@email.ncu.edu.cn}
}

\begin{document}
\maketitle
\begin{abstract}

Parametric Computer-Aided Design (CAD) is crucial in industrial applications, yet existing approaches often struggle to generate long sequence parametric commands due to complex CAD models' geometric and topological constraints. To address this challenge, we propose MamTiff-CAD, a novel CAD parametric command sequences generation framework that leverages a Transformer-based diffusion model for multi-scale latent representations. Specifically, we design a novel autoencoder that integrates Mamba+ and Transformer, to transfer parameterized CAD sequences into latent representations. The Mamba+ block incorporates a forget gate mechanism to effectively capture long-range dependencies. The non-autoregressive Transformer decoder reconstructs the latent representations. A diffusion model based on multi-scale Transformer is then trained on these latent embeddings to learn the distribution of long sequence commands. In addition, we also construct a dataset that consists of long parametric sequences, which is up to 256 commands for a single CAD model.  Experiments demonstrate that MamTiff-CAD achieves state-of-the-art performance on both reconstruction and generation tasks, confirming its effectiveness for long sequence (60-256) CAD model generation.

\end{abstract}    
\section{Introduction}
\label{sec:intro}

\begin{figure}[t]
    \centering
    \includegraphics[width=1\linewidth]{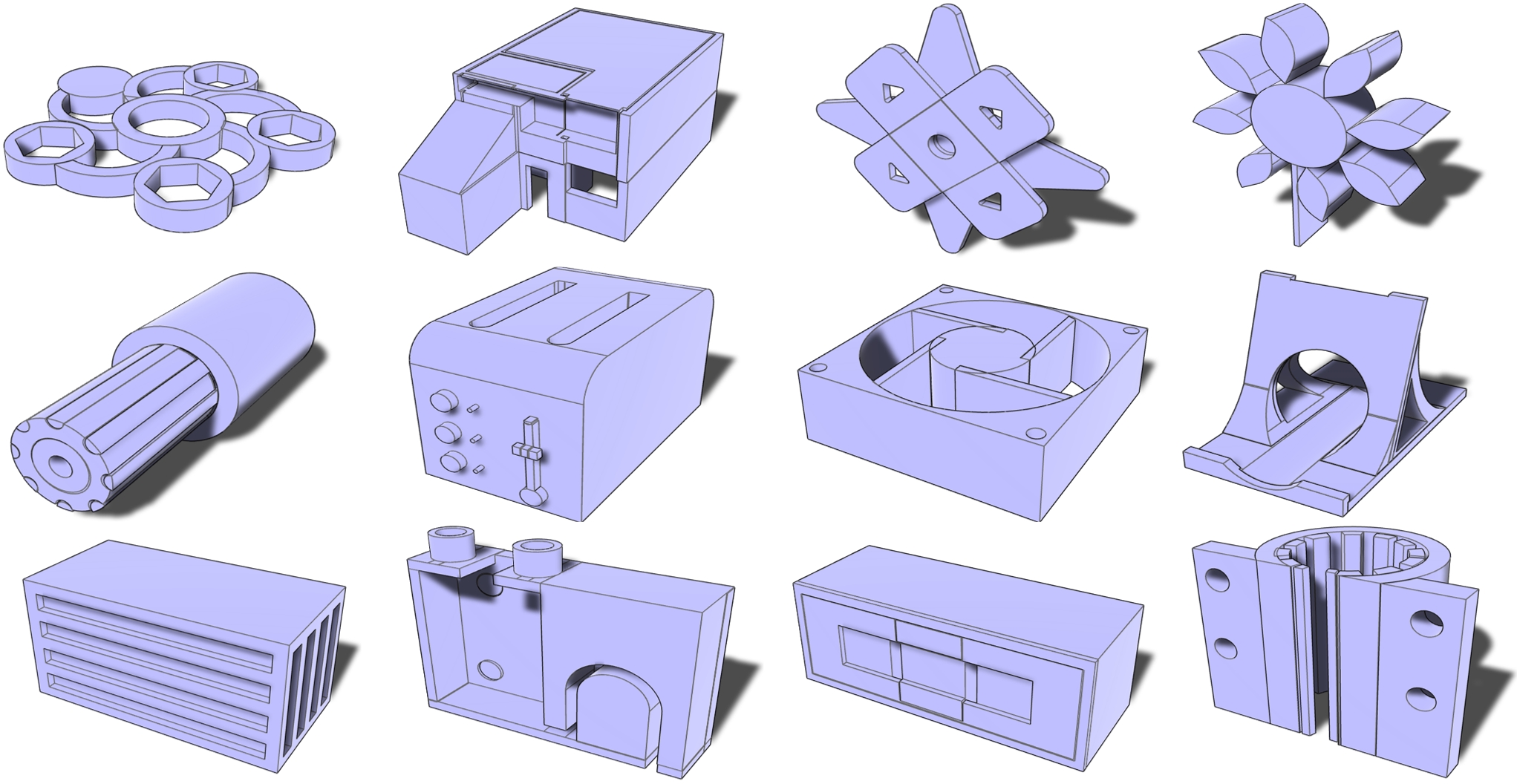}
    \caption{Gallery of Generated CAD Designs. Our generative model infers parametric CAD command sequences, enabling the creation of diverse and structurally valid CAD models. The resulting 3D shapes exhibit clean geometry, well-defined features, and full editability, allowing users to modify designs seamlessly.}
    \label{fig:MamTiff-CAD-gallery}
\end{figure}
Computer-Aided Design (CAD) is a core tool in modern industrial product manufacturing, constructing complex 3D models through parametric command sequences (such as Sketch, Extrusion, and Boolean operations). It not only defines geometric shapes but also thoroughly records design logic and engineering intentions. With the growing demand for highly complex and long sequence CAD models in industrial design, existing generation methods are limited by computational efficiency and sequence modeling capabilities, making it difficult to effectively handle industrial-grade design processes that contain hundreds of commands. 

The key challenges lie in overcoming the bottleneck of long sequence modeling and enabling the automated generation of complex CAD models.
Significant progress has been made in deep learning-based parametric CAD modeling, with Transformer architectures being employed to genarate parametric CAD models~\cite{wu2021deepcad, xu2022skexgen, xu2023hierarchical}. While these approaches achieve strong performance in short-sequence command generation, the inherent quadratic complexity of the Transformer’s self-attention mechanism limits its scalability to longer sequences~\cite{vaswani2017attention}. This constraint makes it difficult to model intricate CAD structures, significantly restricting the applicability of existing methods for generating complex CAD models.

To address these challenges, we propose MamTiff-CAD, a novel CAD command sequence generation framework that integrates Mamba+ and Transformer autoencoder with a multi-scale diffusion model.
In the autoencoding stage, the Mamba+ encoder optimizes state transitions through a gating mechanism, improving its ability to model long-range dependencies. It works with a non-autoregressive Transformer decoder to reconstruct latent representations of CAD command sequences. Once trained, the encoder maps parametric CAD data into a compact, low-dimensional latent space.
In the generation stage, a multi-scale Transformer-driven diffusion model is employed. Through hierarchical denoising, it synchronizes local geometric details with global topological constraints, generating logically coherent latent vectors from Gaussian noise. These vectors are then decoded into executable parametric CAD command sequences.
We make following contributions in this paper:
\begin{itemize}

    \item proposing a novel autoencoder architecture that combines Mamba+ and Transformer to model long sequence parametric CAD commands.

    \item proposing a multi-scale diffusion-based CAD sequence generator to fuse local and global topology, improving both generation quality and 
    \item evaluations showing that our approach achieving state-of-the-art (SOTA) performance on a long sequence CAD dataset containing 13,705 samples with sequence lengths ranging from 60 to 256. 
\end{itemize}

\section{Related Work}
\label{sec:formatting}

\subsection{CAD Model Generation}
Parametric CAD generation has recently shifted from traditional geometric representations to deep learning approaches emphasizing sequence modeling. DeepCAD~\cite{wu2021deepcad} pioneered the use of a Transformer architecture for autoregressive CAD command generation, but it was constrained to short sequences (length $<$ 60), limiting its ability to capture hierarchical structures in complex models. Follow-up efforts introduced hierarchical representations to improve generation accuracy. SkexGen~\cite{xu2022skexgen} proposed a decoupled approach for sketching and extrusion operations, facilitating geometric-topological separation, while HNC-CAD~\cite{xu2023hierarchical} built independent codebooks for rings, sketches, and extrusions via a VQ-VAE~\cite{van2017neural} framework, enabling multi-granularity control. Nevertheless, these methods still grapple with the quadratic computational overhead of Transformer-based architectures, hindering their ability to generate complex, long-sequence CAD models.

The rise of multimodal generation techniques has opened new pathways for CAD generation. CAD-MLLM~\cite{xu2024cad} constructed the first multimodal dataset supporting text, image, and point cloud inputs. However, its generative model still faces topological rupture issues in complex geometric overlapping regions. FlexCAD~\cite{zhang2024flexcad}, by fine-tuning LLMs\cite{hu2022lora}, has achieved controllable CAD generation, but the generated models are relatively simple and do not meet the complex design demands of industrial-grade applications.

\subsection{Long-Sequence Modeling Techniques}
A key challenge in long-sequence modeling is balancing computational efficiency with long-term dependency capture.
Traditional methods~\cite{elman1990finding, 6795963} suffer from the problem of vanishing gradients, making it difficult to effectively capture long-term dependencies. 
While Transformers address this via self-attention, their quadratic complexity in sequence length drives up memory and training costs.
To address this issue, sparse attention mechanisms~\cite{mei2021image, beltagy2020longformer, kitaev2020reformer, wang2020linformer} and local window strategies~\cite{liu2021swin, pan2021mrk, ren2022beyond}have been proposed. These methods can alleviate computational bottlenecks to some extent. However, they sacrifice global context awareness, which can lead to design logic breakdowns and information loss when processing CAD sequence generation.

State Space Models~\cite{hamilton1994state} offer a novel approach for long-sequence modeling. Mamba~\cite{gu2023mamba} incorporates a selective scanning mechanism to achieve long-sequence modeling with linear complexity, demonstrating significant advancements in fields such as audio~\cite{wang2023tf, li2024spmamba, chao2024investigation}, vision~\cite{liu2024vision, liu2025vmamba, cui2024loma, li2024videomamba}, and 3D~\cite{liu2024point, wang2024pointramba}. However, Mamba's application in the CAD domain remains underexplored. In this work, we adapt Mamba for CAD sequence modeling and refine its structure to better capture long-range dependencies.

\begin{figure*}[t]
    \centering
    \includegraphics[width=1\linewidth]{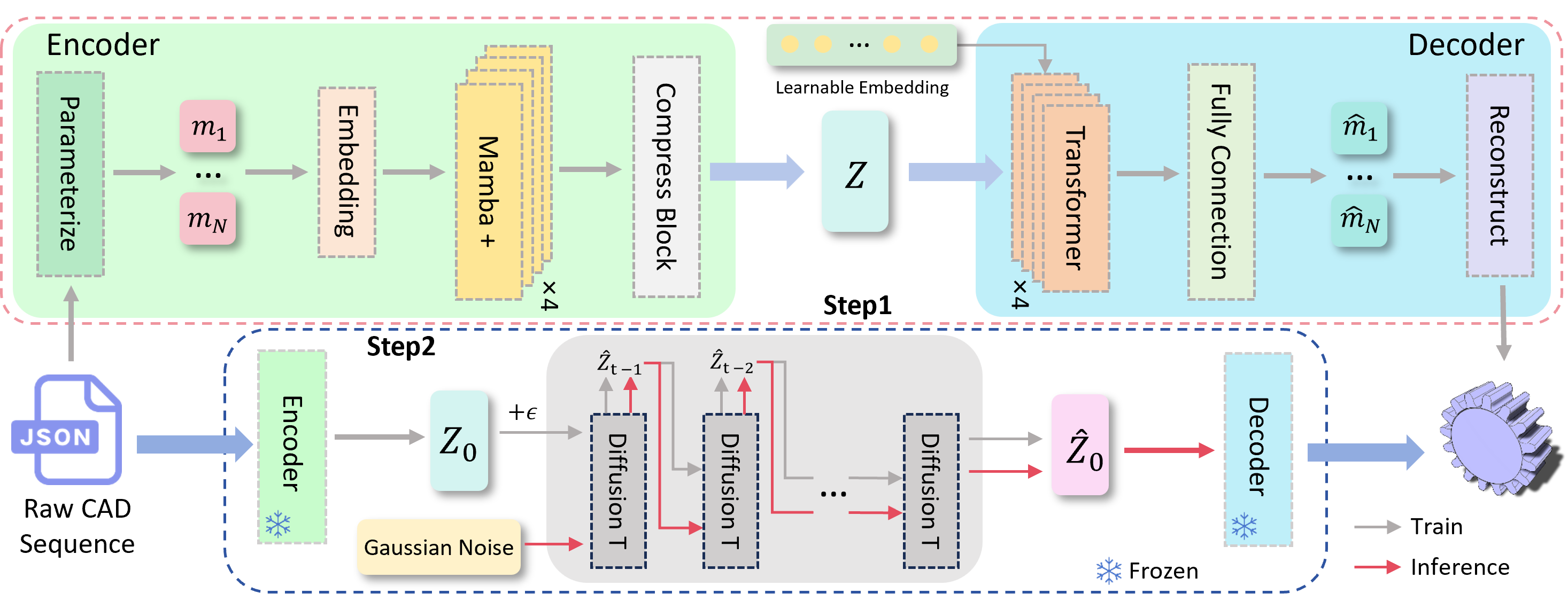} 
    \caption{
    The framework of MamTiff-CAD consists of two main steps. In Step 1, an autoencoder integrating Mamba and Transformer architectures is used to learn the latent representation of CAD command sequences. In Step 2, a diffusion-based module is employed to model the generative distribution of the learned latent representations. During inference, the model directly generates latent variables, which are subsequently decoded to reconstruct the CAD command sequences.
    }
    \label{fig:mamdiff-architecture}
\end{figure*}

\subsection{3D Generation with Diffusion Models}
Diffusion models, as an advanced generative approach, achieve high-quality generation through the process of gradually adding Gaussian noise to the data and then performing denoising in reverse \cite{ho2020denoising}. In the field of 3D generation, diffusion models have been widely applied to tasks such as point cloud to mesh \cite{gupta20233dgen, lyu2023controllable} image to 3D reconstruction \cite{tyszkiewicz2023gecco, liu2023meshdiffusion, xu2024instantmesh, ma2023learning} and CAD model generation \cite{xu2024brepgen,gao2024diffcad}.

Recent studies have made significant progress in combining diffusion models with 3D generation. Generating 3D models through point clouds remains the mainstream method \cite{luo2021diffusion, vahdat2022lion, zhou20213d}. For example, 3DShape2VecSet\cite{zhang20233dshape2vecset} proposed a latent set diffusion framework, achieving 3D shape generation and point cloud completion through multi-scale feature fusion. DiT-3D\cite{mo2023dit} is the first Diffusion Transformer architecture designed for 3D shape generation, leveraging the denoising process of DDPM on voxelized point clouds to efficiently generate high-fidelity 3D models. DiffCAD \cite{gao2024diffcad} employs diffusion models to learn the probabilistic model of CAD generation and alignment. GetMesh \cite{lyu2024getmesh} uses a latent set diffusion model to generate meshes from point clouds, while Polydiff \cite{alliegro2023polydiff} uses a discrete denoising diffusion model to process mesh data, generating 3D triangular mesh structures. Diffusion-SDF \cite{li2023diffusion} generates 3D shapes via diffusion models, offering flexibility in generating diverse shapes that match textual descriptions.

Although these methods have shown potential in generating complex 3D structures, they still face challenges when dealing with parametric CAD sequences. Existing methods primarily focus on generating geometric shapes, neglecting the design logic and parametric constraints of CAD models, which makes the generated results difficult to directly apply to industrial design. The MamTiff-CAD framework proposed in this paper is the first to introduce diffusion models into the field of parametric CAD sequence generation. By learning in latent space vectors and combining Mamba's efficient long-sequence modeling capabilities, it significantly enhances the stability of long-sequence generation.



\section{Method}

\subsection{Architecture Overview}
As shown in Figure~\ref{fig:mamdiff-architecture}, our MamTiff-CAD architecture consists of two steps. In the first step, we train a autoencoder to encode CAD command sequences into latent representation. The input CAD command sequences is represented as a set of parameterized commands \( M = \{m_i\}_{i=1}^{N_c} \), where $N_c$ is the fixed sequence length set to 256. This CAD first passes through an embedding layer and is then fed into Mamba+ blocks. Afterwards, it undergoes compression via a Compress Block for the convenience of following training, generating a latent vector \( Z \). The Transformer-based decoder takes both a learnable embedding and \( Z \) as inputs and subsequently outputs the predicted command sequence \( \hat{M} = \{\hat{m}_i\}_{i=1}^{N_c} \). In the second step, we use diffusion-based networks to learn the generation of encoded latent representation \( Z \). During inference, MamTiff-CAD uses the trained diffusion network to generate the latent representation $\hat{Z}$, which is then decoded into a command sequence using the frozen decoder.

\subsection{AutoEncoder for Latent Representation}

\textbf{Parametric CAD Representation.} 
We follow DeepCAD~\cite{wu2021deepcad} to represent each CAD model as a sequence of parametric commands \( M = \{(C_1, p_1), \dots, (C_n, p_n)\} \), where \( C_i \) is a command type and \( p_i \) is its associated parameter set. For neural network input, we set the sequence length to 256 and end with the special token \(\langle \text{EOS} \rangle\). For sequences less than 256, we pad the length to 256. Each command’s parameters \( p_i \) are encoded as a 16-dimensional vector\footnote{See Supplementary Materials for details.}:
\( \mathbf{p}_i = \begin{bmatrix} 
x, y, \alpha, f, r, \theta, \phi, \gamma, p_x, p_y, p_z, s, e_1, e_2, b, u 
\end{bmatrix} \in \mathbb{R}^{16} \),where unused entries are padded with -1. Continuous parameters are normalized to a \( 2 \times 2 \times 2 \) cube and quantized into 256 discrete levels to preserve geometric consistency. All parameters are eventually converted to discrete tokens for learning.

\begin{figure}[t]
    \centering
    \includegraphics[width=1\linewidth]{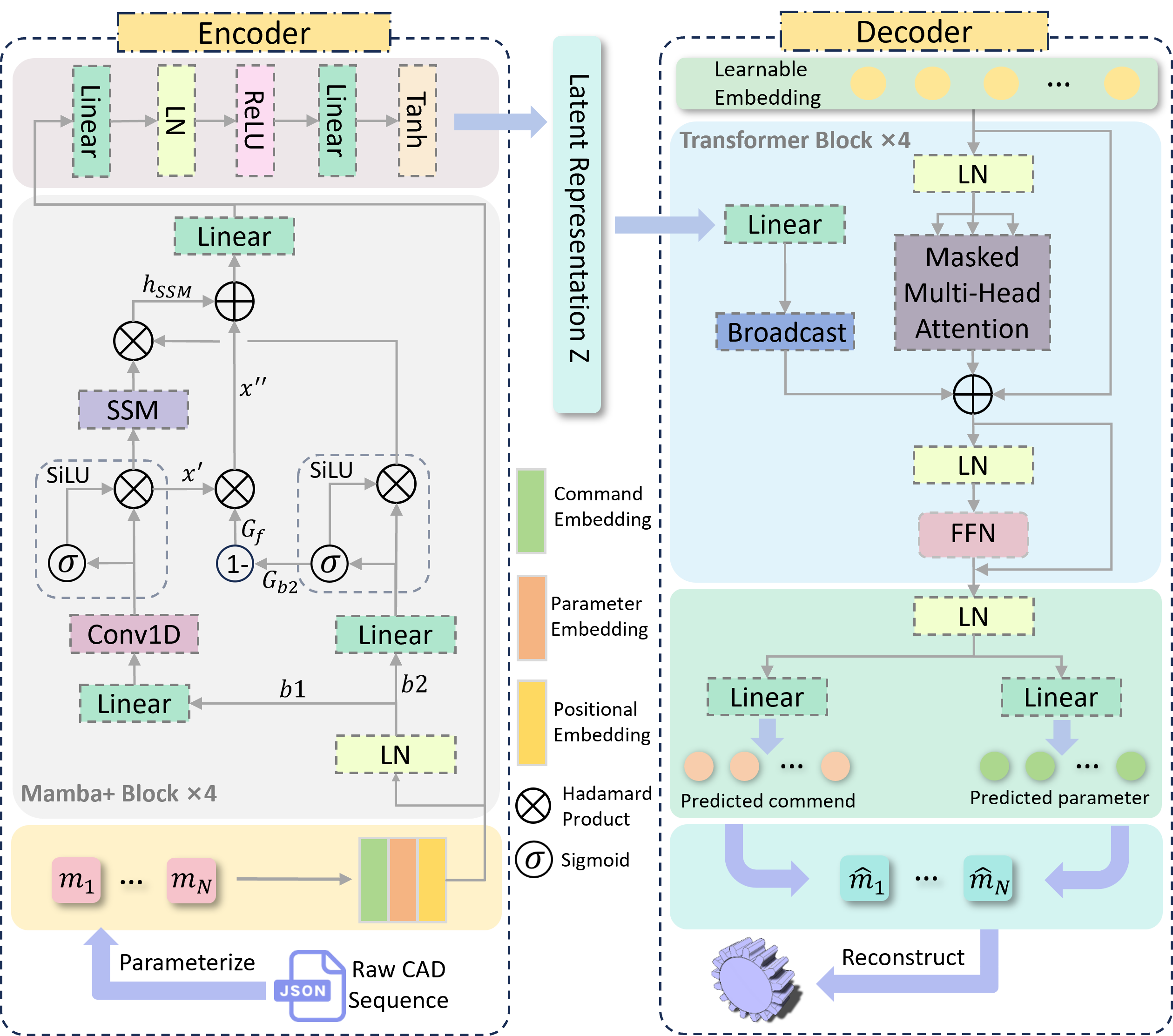}
    \caption{An overview of our autoencoder architecture. The input CAD command sequence is first parameterized and fused with command, parameter, and positional embeddings. It is then processed through four Mamba+ blocks and mapped to the latent representation \( Z \) via a compression bolck. Finally, the Transformer decoder reconstructs the CAD sequence, predicting command types and their corresponding parameters.}
    \label{fig:Autoencoder_architecture}
\end{figure}

\noindent \textbf{Forget Gate Encoder.} To encode CAD commands into the latent space \( Z \), we employ a Forget Gate Encoder. We first embed the parametric representation by considering the command type $C_i$, parameters $p_i$, and the command’s position within the sequence $pos_i$. The embedding of each CAD command, $\text{Emb}(m_i)$, is defined as follows:
\begin{equation}
    \text{Emb}(m_i)= W_{\text{cmd}} \delta_{C_i} + W_{\text{pa}} \cdot \text{flatten}(W_{\text{pb}} \delta_{p_i}) + W_{\text{pos}} \delta_i
\end{equation}

Here, $W_{\text{cmd}} \in \mathbb{R}^{d_E \times 6}$ is a learnable matrix and $\delta_{C_i} \in \mathbb{R}^{6}$ is a one-hot vector indicating the command type. The system defines six command types. Each CAD command comprises 16 parameters, each quantized into an 8-bit integer. We convert each integer into a one-hot vector of dimension 257, where 256 dimensions correspond to the quantized values and the remaining one dimension indicates that the parameter is unused. Stacking these 16 one-hot vectors forms a matrix $\delta_{p_i} \in \mathbb{R}^{257 \times 16}$. Then, using the learnable matrix $W_{\text{pb}} \in \mathbb{R}^{d_E \times 257}$, we embed each parameter individually, flatten the resulting embeddings, and apply a linear transformation via $W_{\text{pa}} \in \mathbb{R}^{d_E \times 16d_E}$. Finally, $W_{\text{pos}} \in \mathbb{R}^{d_E \times N_c}$ is another learnable matrix, and $\delta_i \in \mathbb{R}^{N_c}$ is a one-hot vector that marks the command’s position within the sequence, with $N_c=256$ being the maximum sequence length.

We then use four Mamba+ blocks, i.e. optimized SSM-based modules integrated with a forget gate mechanism, as our main encoder block. Specifically, the Mamba+ block employs a dual-branch structure for information processing. The feature transformation branch ($b1$) extracts sequence features using 1D convolution and SSM blocks, while the gating mechanism branch ($b2$) generates control signals for information flow using the SiLU activation function. To further enhance the retention of historical information, we introduce a forget gate $\text{G}_f$ as:
\begin{equation}
    \text{G}_f = 1 - \text{G}_{b2}
\end{equation}

\( \text{G}_{b2} \) is calculated using the Sigmoid function, controlling the flow of current input information. The forget gate \( \text{G}_f \), modulates the output \( x' \) from $b1$, ensuring that some historical information is not completely discarded. The updated historical information is given by $x'' = \text{G}_f \cdot x'$.
The final output of Mamba+ blocks integrates the SSM computation result with the feature adjusted by the forget gate $h_{\text{out}} = x'' + h_{\text{SSM}}$. This design enables the Mamba+ block to effectively extract new features while preserving long-term dependencies, enhancing its ability to process complex CAD command sequences efficiently. 

To efficiently encode long sequence CAD commands, we construct the Mamba+ block by stacking four layers of Mamba+ blocks, capable of modeling CAD command sequences ranging from 60 to 256 in length. The encoder first converts the input command sequence \( M = \{m_1, m_2, \dots, m_{256}\} \) into continuous representations through an embedding layer, then progressively captures local and global dependencies through hierarchical processing in the Mamba+ blocks.

\noindent \textbf{Decoder.} We employ a Transformer-based decoder, which consists of four blocks, to reconstruct parametric CAD command sequences. The decoder's input comprises two parts: 1) a latent vector \( Z \), and 2) a set of learnable embedding vectors produced by a learnable embedding layer, which provides a representation for each position in the sequence. In our model, the maximum sequence length is set to \( N_c = 256 \). The decoder processes these inputs layer by layer to capture global dependencies among commands in long sequences and ultimately generates the predicted CAD command sequence \( \hat{M} = \{\hat{m}_1, \hat{m}_2, \dots, \hat{m}_{256}\} \), where each command \( \hat{m}_i \) consists of a command type \( \hat{C}_i \) and parameters \( \hat{\mathbf{p}}_i \). Our objective is to maximize the generation probability:
\begin{equation}
p(\hat{M} \mid z, \Theta) = \prod_{i=1}^{N_{c}} p(\hat{C}_i, \hat{\mathbf{p}}_i \mid z, \Theta)
\end{equation}

Here, \( \Theta \) denotes the decoder's network parameters. Finally, the decoder's output hidden states are mapped through a linear transformation layer into the specific command type and parameter space, thereby completing the generation of the entire CAD command sequence.

\noindent \textbf{Training of the Autoencoder.} We train our autoencoder by reconstructing parameterized CAD command sequences to learn the latent representation \( Z \) of CAD models. The training objective is to minimize the discrepancy between the predicted outputs and the ground truth, ensuring that the model accurately captures the structural information of the parameterized CAD sequences. We employ a cross-entropy loss function to supervise the learning of both CAD command types and parameters.

Our output sequence is fixed to \(N_c = 256\) commands, each with \(N_p = 16\) parameters. For the \(i\)-th command, the ground truth command type is denoted by \(t_i\) (derived from the true command \(C_i\)), and the corresponding ground truth parameters are represented as \(\{a_{i,j}\}_{j=1}^{N_p}\) (where each \(a_{i,j}\) is an element of the previously defined parameter vector). The model predicts a probability distribution over command types \(p_i\) such that \(p_i(t_i)\) represents the probability assigned to the true command type, and for each parameter \(a_{i,j}\), it predicts a probability distribution \(q_{i,j}\) over quantized parameter values.The total loss is defined as:
\begin{equation}
L = \sum_{i=1}^{N_{c}} \ell\big(p_i(t_i)\big) + \beta \sum_{i=1}^{N_{c}} \sum_{j=1}^{N_{p}} \ell\big(q_{i,j}(a_{i,j})\big)
\end{equation}
where \(\beta\) is set to 2 to balance the parameter loss against the command type loss. Contributions from padding commands, such as the \(\langle \text{EOS} \rangle\) token, and unused parameters labeled as \(-1\) are ignored during loss computation. For further details on the experimental configurations, please refer to the Appendix.

\subsection{Diffusion Generator}

\begin{figure}[t]
    \centering
    \includegraphics[width=1\linewidth]{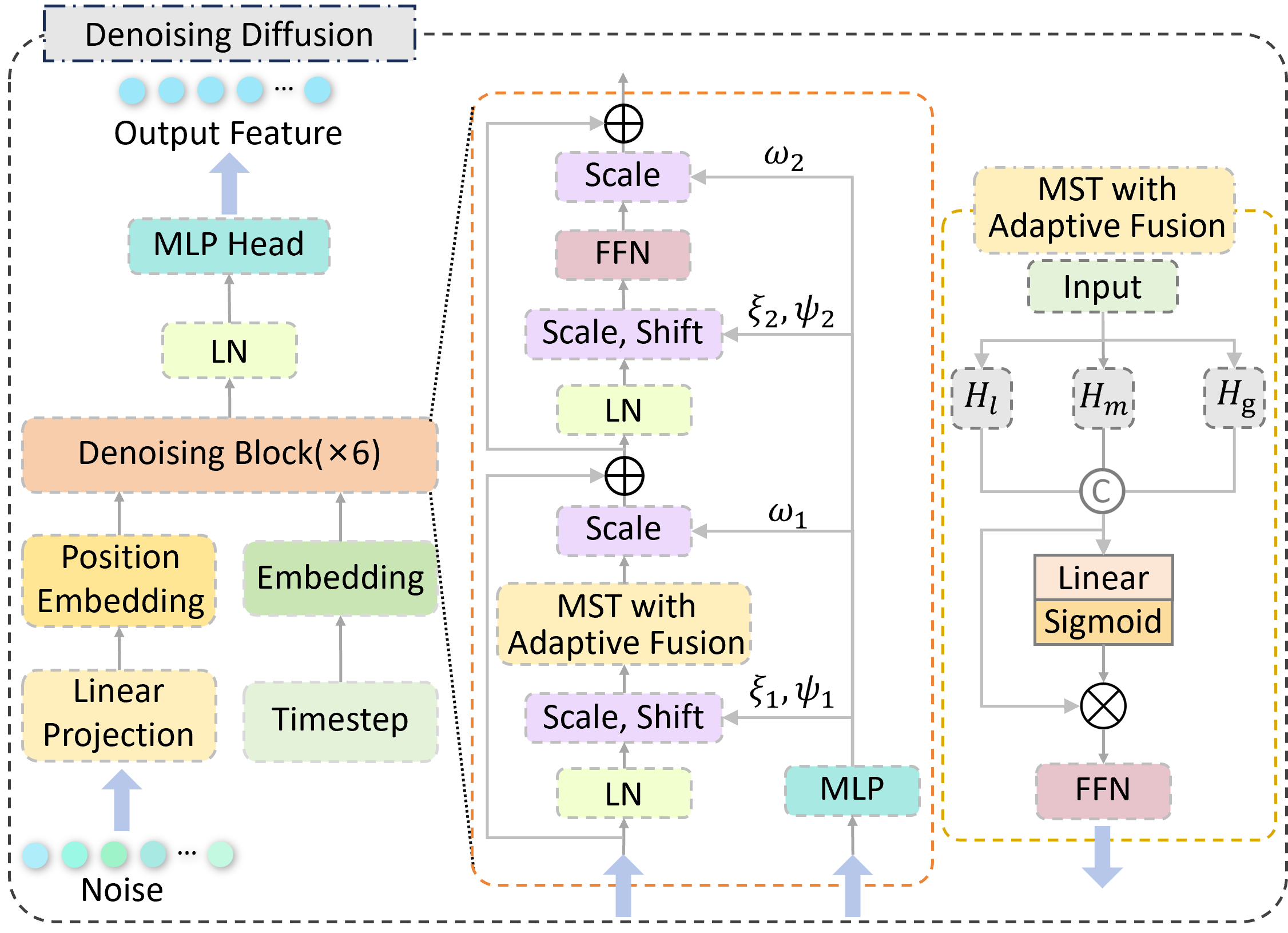}
    \caption{Our denoising diffusion model. The input noise undergoes linear projection with positional and timestep embeddings, followed by feature extraction through encoder blocks (×6). The core denoising structure, MST with Adaptive Fusion, integrates multi-scale attention and adaptive fusion to dynamically adjust feature distributions. Finally, the MLP Head reconstructs the denoised output features.}
    \label{fig:diffusion-structure}
\end{figure}

The above frozen autoencoder encodes the command sequences into a compact latent variable \(Z \in \mathbb{R}^{N \times 256 \times 64}\). In the forward process, we employ a linear variance schedule \(\{\beta_t\}\) to add noise into the latent space, while the denoising process is guided by a Multi-Scale Transformer-based Diffusion Generator (MST-D), as shown in Figure~\ref{fig:diffusion-structure}.


\noindent \textbf{Multi-Scale Transformer Denoiser.}
The denoising network is composed of stacked multi-scale Transformer layers, each following a three-stage pipeline: \textit{Hierarchical Attention}: Each layer contains three parallel attention branches with window sizes of 64, 128 and 256 to capture local geometric constraints, medium-range topological dependencies, and global semantic coherence respectively, processing the latent sequence at three different scales. \textit{Adaptive Fusion}: The outputs of these attention branches are concatenated and fused via a learnable gating mechanism:
    \begin{equation}
        \mathbf{H} = \text{MLP} \left( \sigma(\mathbf{W}_g [\mathbf{H}_l \| \mathbf{H}_m \| \mathbf{H}_g]) \odot [\mathbf{H}_l \| \mathbf{H}_m \| \mathbf{H}_g] \right)
    \end{equation}
where \(\sigma(\cdot)\) is the sigmoid function, and \(\|\) denotes concatenation.
\textit{Time-Step Embedding}: This module inputs temporal information into each Transformer block by transforming the diffusion time step \( t \) into an embedding vector. 
The embedding is then processed through an MLP to generate two sets of scale and shift parameters $\{ \xi_i, \psi_i, \omega_i \}$. Here, $\xi_i$ and $\psi_i$ modulate the layer-normalized activations, while $\omega_i$ adjusts the strength of the residual connections.
Together, these parameters enable the Transformer to adapt its behavior in a time-aware manner.

\noindent \textbf{Sequence-Aware Positional Encoding.}  
To preserve the sequential logic of CAD commands, we employ a scalable sinusoidal positional encoding \(PE\), defined as
\begin{equation}
PE(Z) = Z + \eta \cdot \text{PE}(\text{pos}, D)
\end{equation}
where \(\eta \in \mathbb{R}\) is a trainable scalar that adaptively adjusts the positional weighting. Instead of writing the full vector in one long equation, we define the positional encoding function piecewise as follows:
\begin{equation}
\begin{split}
\text{PE}_{2i}(\text{pos}, D) &= \sin\Bigl(\frac{\text{pos}}{10000^{2i/D}}\Bigr)\\[1ex]
\text{PE}_{2i+1}(\text{pos}, D) &= \cos\Bigl(\frac{\text{pos}}{10000^{2i/D}}\Bigr)
\end{split}
\end{equation}
for \(i = 0,1,\dots,\frac{D}{2}-1\).  
Here, \(\text{pos}\) denotes the sequence position, and \(D\) is the embedding dimension. This encoding scheme provides both absolute and relative position information, enabling the model to capture the sequential dependencies of CAD commands more effectively.

\noindent \textbf{Training and Generation.}
The denoising network learns to predict the noise injected at each timestep, modeling the distribution of the target latent space. Given a clean latent vector \(Z_0\) from the frozen encoder, we first sample a timestep \(t \sim U(1, T)\) and add noise according to the linear variance schedule \(\{\beta_t\}\):
\begin{equation}
Z_t = \alpha_t Z_0 + \sqrt{1 - \alpha_t}\,\epsilon
\end{equation}
where \(\epsilon \sim \mathcal{N}(0, I)\), \(\alpha_t = \prod_{s=1}^{t}(1 - \beta_s)\) controls the rate of noise accumulation. The network \(\epsilon_{\theta}\) is then trained to predict \(\epsilon\) by minimizing the mean squared error:
\begin{equation}
L_{\text{diff}} 
= \mathbb{E}_{t, Z_0, \epsilon}\bigl[\|\epsilon - \epsilon_{\theta}(Z_t, t)\|_2^2\bigr]
\end{equation}
\textit{Sampling Process:} Starting from Gaussian noise \(Z_T\), the latent vector is iteratively denoised using:
\begin{equation}
Z_{t-1} = \frac{1}{\alpha_t}\Bigl(
Z_t 
- \beta_t \frac{1}{\sqrt{1 - \alpha_t}}
\,\epsilon_{\theta}(Z_t, t)
\Bigr) 
+ \beta_t z
\end{equation}
where \(z \sim \mathcal{N}(0, I)\). Here, \(\alpha_t\) and \(\beta_t\) denote the attenuation factor and variance schedule parameter at timestep \(t\), respectively. Through this process, our diffusion generator progressively removes noise in the latent space to yield CAD command sequences with global consistency and rich structural information.




\begin{table}[t]
\caption{Statistical comparison between the DeepCAD dataset and our ABC-256 dataset, including the total number, the average length of parametric CAD sequences, and the length distribution.}
  \centering
  \resizebox{\columnwidth}{!}{ 
    \begin{tabular}{@{}lcccccc@{}}
      \toprule
      Dataset & Total & Average Length & 1-10 & 11-60 & 61-128 & 129-256 \\
      \midrule
      DeepCAD & 178,238 & 15 & 44.58 & 55.42 & - & - \\
      ABC-256 & 13,705 & 99 & - & - & 82.89 & 17.11 \\
      \bottomrule
    \end{tabular}
  }
  \label{tab:dataset}
\end{table}

\section{ABC-256 Dataset}
Although several publicly available CAD datasets exist, such as ABC~\cite{koch2019abc}, DeepCAD~\cite{wu2021deepcad}, Fusion 360 Gallery~\cite{willis2021fusion}, and CAD as Language~\cite{ganin2021computer}, datasets that fully capture the CAD design process remain scarce. The ABC dataset contains 1 million B-rep models from Onshape~\cite{Onshape}, but lacks design operation sequences, making full process reconstruction difficult. The Fusion 360 Gallery dataset provides CAD command sequences based on sketch profile extrusions, but these sequences are short, limiting their suitability for long sequence CAD generation. DeepCAD includes 178,238 CAD models with parameterized sequences, but its maximum sequence length of 60 constrains its ability to model complex, long sequence CAD designs.

To address this, we constructed and publicly released a complex CAD command sequence dataset to advance research in this field. Using ABC dataset model links, we leveraged the Onshape API and FeatureScript to extract CAD commands from CSG representations and convert them into parameterized sequences. We then filtered models to retain only those with complete operations, excluding simpler cases with only sketching and extrusion, ensuring sequence lengths between 60 and 256.
The final dataset includes 13,705 CAD models, split into 10,964 for training, 1,370 for validation, and 1,371 for testing. Compared to DeepCAD, our dataset has a 6.6× longer average sequence length (Table~\ref{tab:dataset}), offering greater complexity in both length and distribution, making it more suitable for long sequence CAD model generation.

\begin{figure*}[t]
    \centering
    \includegraphics[width=1\linewidth]{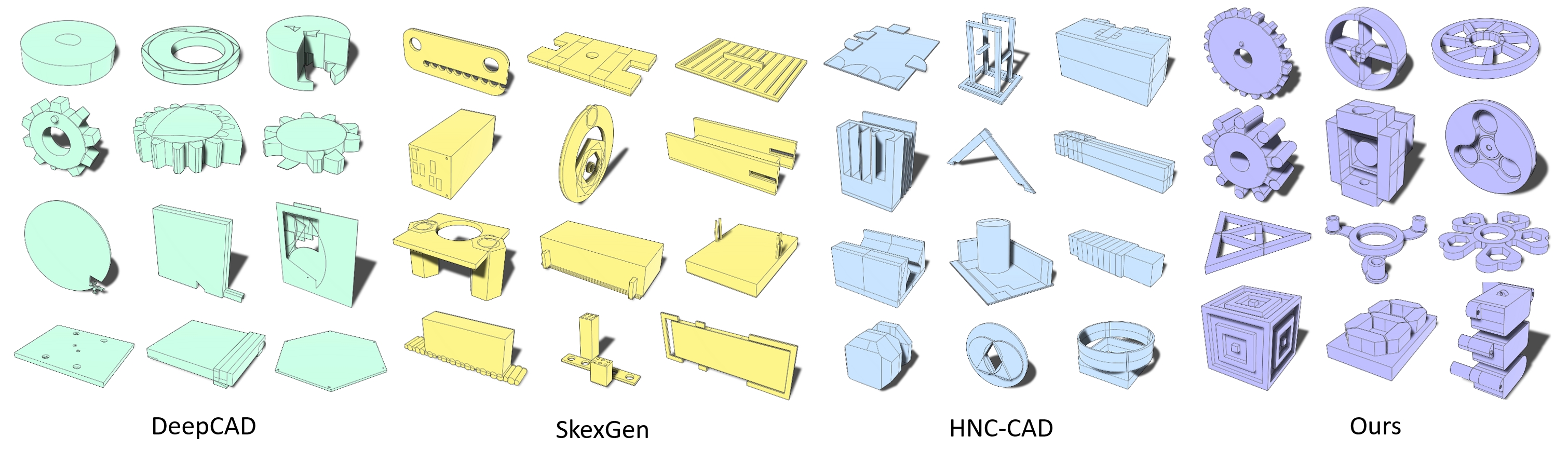} 
    \caption{This figure compares the CAD models generated by DeepCAD, SkexGen, HNC-CAD, and MamTiff-CAD, highlighting differences in shape complexity and geometric details among the methods. It can be observed that MamTiff-CAD produces models with higher structural integrity and greater complexity.} 
    \label{fig:all_generate} 
\end{figure*}

\section{Experiment}
\subsection{Implementation details}

We implemented the MamTiff-CAD model based on the PyTorch framework and completed the training of the model on an NVIDIA RTX 4090 GPU.
During training, we adopted the AdamW optimizer \cite{loshchilov2017fixing} with an initial learning rate of 0.001, applying linear warmup for the first 2000 steps. Additionally, we introduced gradient clipping to prevent gradient explosion issues. The training of MamTiff-CAD is divided into two stages. The first stage involves training for 300 epochs with a batch size of 32. In the second stage, training for 200,000 epochs with a batch size of 64 to ensure the model has enough time to fully learn and reach convergence. 

\subsection{Evaluation Metrics}
We comprehensively evaluate the performance of the autoencoder using five metrics, including \textit{Command Accuracy}, \textit{Parameter Accuracy}, \textit{Median Chamfer Distance}, \textit{Invalid Ratio}, and \textit{Step Ratio}. 

\begin{itemize}
    \item \textit{Command Accuracy} (ACC$_{\text{c}}$): the match between the predicted CAD command type and the ground truth.
    \item \textit{Parameter Accuracy} (ACC$_{\text{p}}$): the match of command parameters under the condition that the command type is correctly predicted.
    \item \textit{Median Chamfer Distance} (MCD): the geometric error between the reconstructed 3D shape and the reference shape, computed as the median distance between corresponding points.
    \item \textit{Invalid Ratio} (IR): the proportion of CAD models that cannot be converted into point clouds.
    \item \textit{Step Ratio} (SR): the proportion of generated CAD sequences successfully converted to the STEP format.
\end{itemize}

For the performance evaluation of the generator, we adopt six metrics, including \textit{Coverage} (COV), \textit{Jensen-Shannon Divergence} (JSD), \textit{Minimum Matching Distance} (MMD), \textit{Novelty}, \textit{Uniqueness}, and \textit{Step Ratio}. The specific metric descriptions are as follows:

\begin{itemize}
    \item \textit{Coverage} (COV): whether the generated data covers the distribution of real data.
    \item \textit{Jensen-Shannon Divergence} (JSD): the difference between the generated data distribution and the real data distribution.
    \item \textit{Minimum Matching Distance} (MMD): the minimum average distance between generated and real samples.
    \item \textit{Unique}: the proportion of non-duplicate samples in the generated dataset.
    \item \textit{Novel}: the proportion of samples in the generated dataset that do not appear in the training set.
\end{itemize}

\begin{table}[b]
  \caption{Autoencoding performance on ABC-256 dataset. Comparison of MamTiff-CAD and other models in CAD sequence reconstruction. MCD is multiplied by $10^3$, and ACC$_{c}$, ACC$_{p}$, and IR are multiplied by 100\%.}
  \centering
  \resizebox{\columnwidth}{!}{ 
    \begin{tabular}{@{}lccccc@{}}
      \toprule
      Method & ACC$_{c}$ $\uparrow$ & ACC$_{p}$ $\uparrow$ & MCD $\downarrow$ & IR $\downarrow$ & SR $\uparrow$ \\
      \midrule
      DeepCAD & 92.24 & 75.93 & 41.02 & 33.11\% & 70.46\% \\
      MT-CAD & 89.72 & 66.87 & 121.35 & 39.89\% & 63.97\% \\
      MLSTM-CAD & 86.09 & 65.55 & 112.89 & 42.53\% & 59.85\% \\
      \textbf{OURS} & \textbf{99.99} & \textbf{99.93} & \textbf{0.75} & \textbf{8.50\%} & \textbf{93.93\%} \\
      \bottomrule
    \end{tabular}
  }
  \label{tab:autoencoder result}
\end{table}

\subsection{Autoencoding Parametric CAD }

To comprehensively evaluate the effectiveness of our proposed MamTiff-CAD framework, we conducted a comparative analysis across multiple models, including DeepCAD, MT-CAD, MLSTM-CAD, and our own model. Among them, MT-CAD integrates Mamba with Transformer, while MLSTM-CAD combines Mamba with LSTM.
Experimental results confirm that our framework offers significant advantages in long sequence parametric CAD modeling, further reinforcing the rationale behind our design decisions.

\begin{table}[t]
  \caption{Quantitative evaluation of generalization ability on the Fusion360 reconstruction dataset.}
  \centering
  \resizebox{\columnwidth}{!}{ 
    \begin{tabular}{@{}lccccc@{}}
      \toprule
      Method & ACC$_{c}$ $\uparrow$ & ACC$_{p}$ $\uparrow$ & MCD $\downarrow$ & IR $\downarrow$ & SR $\uparrow$ \\
      \midrule
      DeepCAD & 93.35 & 77.99 & 104.76 & 19.41\% & 82.57\% \\
      MT-CAD & 91.85 & 60.18 & 300.21 & 11.81\% & 90.17\% \\
      MLSTM-CAD & 88.42 & 62.13 & 261.80 & 23.39\% & 80.97\% \\
      \textbf{OURS} & \textbf{99.99} & \textbf{97.99} & \textbf{1.44} & \textbf{5.70\%} & \textbf{95.16\%} \\
      \bottomrule
    \end{tabular}
  }
  \label{tab:generalization}
\end{table}

\begin{figure}[t]
    \centering
    \includegraphics[width=1\linewidth]{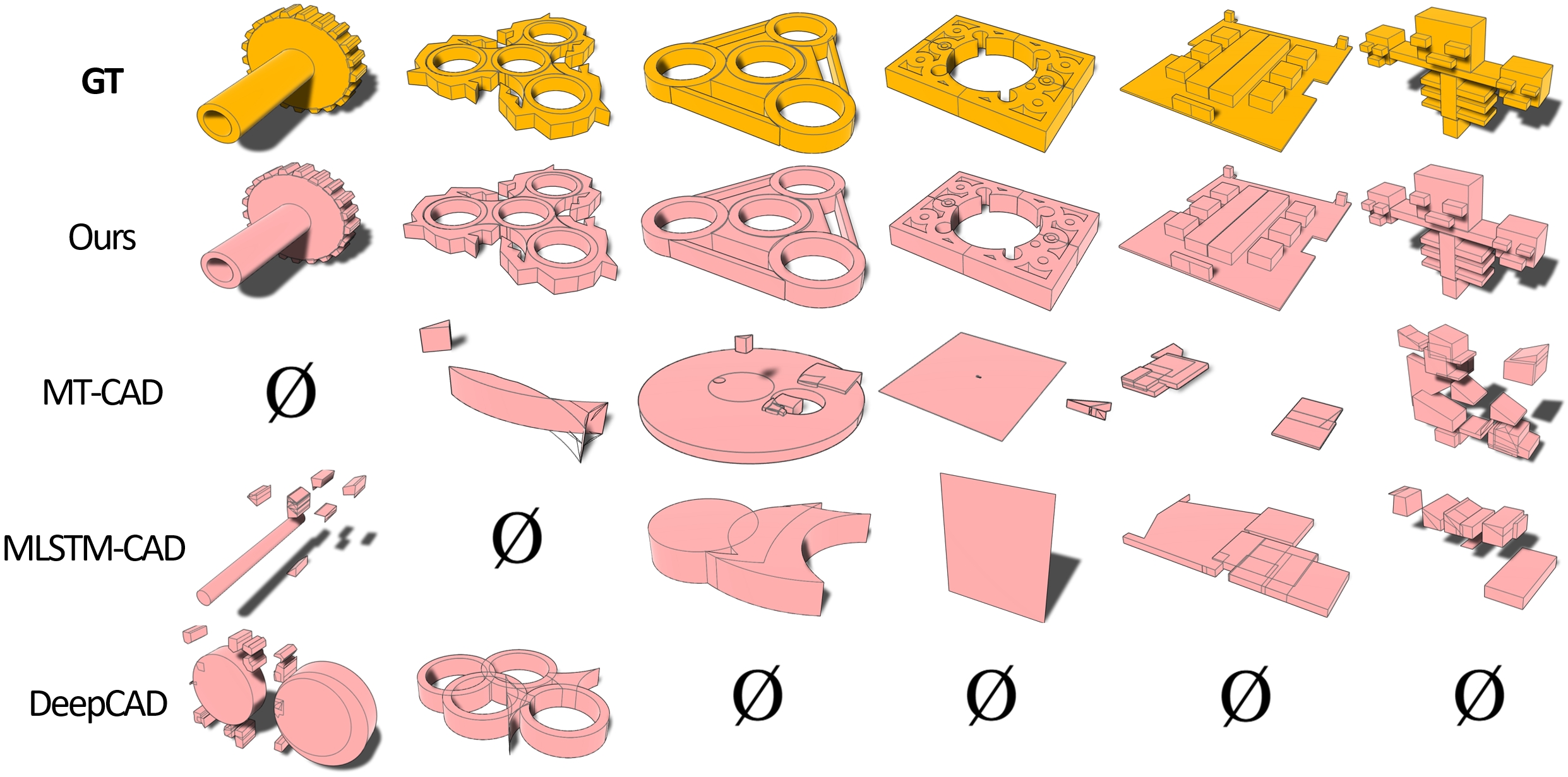}
    \caption{Visual illustrations of CAD sequence reconstruction. GT represents Ground Truth, and Ours refers to MamTiff-CAD. The symbol $\text{\O}$ indicates an invalid CAD sequence, resulting in the failure of 3D shape construction.}
    \label{fig:autoencoder_chongjian}
\end{figure}

\textbf{Results.} The experimental results are presented in Table \ref{tab:autoencoder result}. Overall, our model achieves SOTA performance across all evaluation metrics. Specifically, our proposed MamTiff-CAD attains a command accuracy of 99.99\% and a parameter accuracy of 99.93\%, demonstrating superior capability in long command sequence prediction. In terms of Median Chamfer Distance (MCD), MamTiff-CAD outperforms other baseline models, indicating that the reconstructed 3D shapes are geometrically closer to the reference data. Invalid Ratio and Step Ratio, our model achieves an IR of 8.50\% and a Step Ratio of 93.93\%, indicating a higher proportion of valid command sequences, thereby demonstrating the reliability of MamTiff-CAD. Furthermore, Figure \ref{fig:autoencoder_chongjian} provides a visual comparison of CAD sequence reconstruction results across different models.
MamTiff-CAD produces CAD models with better structural integrity and geometric accuracy, making them closer to the ground truth. This further validates the superior performance of MamTiff-CAD in long sequence parametric CAD prediction.  

\textbf{Generalization Performance.} To evaluate the generalization capability of MamTiff-CAD, we trained the model on the ABC-256 dataset and directly tested it on the Fusion360 dataset, which was not used during training. The Fusion360 dataset consists of approximately 8,000 parametric CAD sequences derived from Autodesk Fusion 360 designs. Although the CAD models in the Fusion360 dataset exhibit simpler structures, which narrows the performance gap between MamTiff-CAD and the baseline models in certain metrics, MamTiff-CAD still consistently outperforms all other models across all evaluation metrics, achieving the best overall performance (see Table \ref{tab:generalization}).

\subsection{Unconditional Generation}

We comprehensively evaluated latent vector-based CAD generation across several models, including DeepCAD, SkexGen, HNC-CAD, and our MamTiff-CAD. 
Each method randomly generated 10,000 samples, which were converted to point clouds for metric-based evaluation. The experimental results in Table~\ref{tab:cad_generation_comparison} indicate that our method achieves state-of-the-art (SOTA) performance in all metrics, particularly excelling in JSD (3.19) and Step Ratio (85.38\%), demonstrating the higher quality and effectiveness of the generated sequences. Furthermore, as shown in Figure \ref{fig:all_generate}, MamTiff-CAD generates more complex CAD models compared to other methods, further validating its advantage in long sequence CAD generation tasks.

\begin{table}[t]
  \caption{The unconditional generation result compared with DeepCAD,  SkexGen, HNC-CAD and ours.}
  \centering
  \resizebox{\columnwidth}{!}{ 
    \begin{tabular}{@{}lcccccc@{}}
      \toprule
      Method & MMD $\downarrow$ & JSD $\downarrow$ & COV $\uparrow$ & Unique $\uparrow$ & Novel $\uparrow$ & SR $\uparrow$ \\
      \midrule
      DeepCAD & 2.66 & 6.49 & 56.66\% & 75.8 & 88.0 & 23.96\% \\
      SkexGen & 2.31 & 4.53 & 57.76\% & 80.5 & 96.9 & 75.26\% \\
      HNC-CAD & 1.63 & 4.25 & 62.03\% & 89.2 & 91.8 & 80.86\% \\
      \textbf{OURS} & \textbf{1.32} & \textbf{3.19} & \textbf{65.31\%} & \textbf{99.6} & \textbf{99.4} & \textbf{85.38\%} \\
      \bottomrule
    \end{tabular}
  }
  \label{tab:cad_generation_comparison}
\end{table}

\subsection{Ablation Studies}  
We conducted two ablation studies to systematically assess the impact of the two key techniques, Mamba+ and Multi-Scale Transformer (MST), on our model's performance.

\textbf{Mamba+ Block.} To investigate the role of the Mamba+ block in reconstructing long sequence parametric CAD commands, we conducted two ablation experiments: (1) replacing the Mamba+ block with Transformer block, and (2) replacing the Mamba+ block with the standard Mamba block. The experimental results (see Table~\ref{tab:mamba vs mamba+}) indicate that, compared to the standard Mamba block, the Mamba+ block achieves improvements across most evaluation metrics. This confirms that the incorporation of the forget gate mechanism enhances the model’s ability to capture long-range dependencies in CAD command sequences, with particularly notable improvements in common CAD sequence tasks over the Transformer-based approach.

\begin{table}[t]
  \caption{Comparison of reconstruction performance between Mamba and Mamba+ block for parametric CAD sequences.}
  \centering
  \resizebox{\columnwidth}{!}{ 
    \begin{tabular}{@{}lccccc@{}}
      \toprule
      Method & ACC$_{c}$ $\uparrow$ & ACC$_{p}$ $\uparrow$ & MCD $\downarrow$ & IR $\downarrow$ & SR $\uparrow$ \\
      \midrule
      Transformer & 77.29 & 63.62 & 64.30 & 67.46\% & 34.28\% \\
      Mamba & 99.98 & 99.90 & 0.76 & 10.35\% & 92.01\% \\
     \textbf{Mamba+} & \textbf{99.99} & \textbf{99.93} & \textbf{0.75} & \textbf{8.50\%} & \textbf{93.93\%} \\
      \bottomrule
    \end{tabular}
  }
  \label{tab:mamba vs mamba+}
\end{table}

\textbf{MST or Non-MST.} To assess the impact of MST on our generative model, we conducted an ablation study by randomly generating 10,000 models and comparing the generation capabilities with and without the MST module. The experimental results in Table~\ref{tab:mst_comparison} demonstrate that incorporating MST significantly improves model performance, particularly in terms of generation quality and validity, confirming the critical role of MST in the diffusion process. 


\begin{table}[t]
  \caption{The unconditional generation result compared between w/o MST and MST.}
  \centering
  \resizebox{\columnwidth}{!}{ 
    \begin{tabular}{@{}lcccccc@{}}
      \toprule
      Method & MMD $\downarrow$ & JSD $\downarrow$ & COV $\uparrow$ & Unique $\uparrow$ & Novel $\uparrow$ & SR $\uparrow$ \\
      \midrule
      w/o MST & 1.47 & 4.92 & 61.69\% & 99.5 & \textbf{99.6} & 77.05\% \\
      \textbf{w/ MST} & \textbf{1.32} & \textbf{3.19} & \textbf{65.31\%} & \textbf{99.6} & 99.4 & \textbf{85.38\%} \\
      \bottomrule
    \end{tabular}
  }
  \label{tab:mst_comparison}
\end{table}

\vspace{-0.5em}
\section{Discussion and Conclusion}

Although MamTiff-CAD has demonstrated exceptional performance in long-sequence parameterized CAD modeling tasks, several limitations remain that warrant further investigation. Firstly, the dataset used in this study does not cover all CAD command types in industrial design, which somewhat limits the model’s ability to handle more complex topological structures. Secondly, this research primarily focuses on the generation and parsing of CAD command sequences and does not directly extract features from Boundary Representation (Brep). Therefore, future research is required to address how to effectively integrate CSG (Constructive Solid Geometry) and Brep representations to enhance the model’s generalization ability and applicability. This remains a challenging but promising direction.

In conclusion, this paper presents MamTiff-CAD, a novel framework that integrates an improved Mamba module with a diffusion model to efficiently model long parameterized CAD command sequences and generate complex 3D shapes. 
We also introduce a dataset of 13,705 CAD models with sequence lengths of 60 to 256, significantly surpassing existing datasets. Experimental results validate MamTiff-CAD’s superiority in sequence reconstruction and generation. Future work will focus on refining the model and exploring its integration with advanced design processes and interactive methods to enhance intelligent CAD design.

\vspace{-0.5em}
\section*{Acknowledgments}
This work was supported by the Fundamental Research Funds for the Central Universities, the Shaanxi Provincial Natural Science Foundation (2024JC-YBQN-0702), the NUS Research Scholarship, the Hunan Natural Science Foundation (2025JJ50338), and the Shanghai Education Committee AI Project (JWAIYB-2).
{
    \small
    \bibliographystyle{ieeenat_fullname}
    \bibliography{main}
}

\clearpage
\setcounter{page}{1}
\maketitlesupplementary
\setcounter{section}{0} 

\section{ABC-256 Dataset}
\begin{figure}[h]
    \centering
    \includegraphics[width=\linewidth]{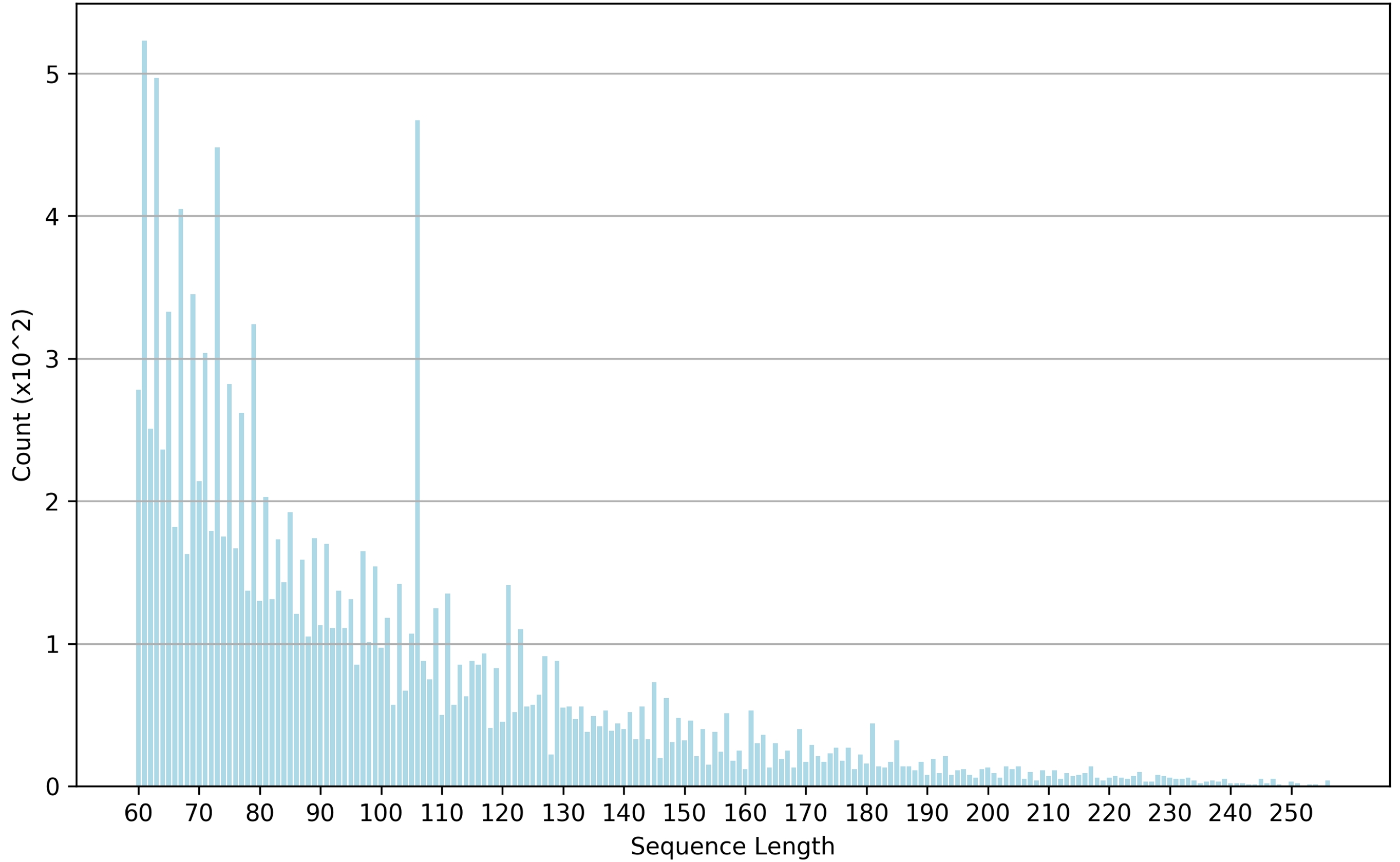}
    \caption{Distribution of CAD command sequence lengths in the ABC-256 dataset.}
    \label{fig:datatongji}
\end{figure}

We use the model links from the ABC~\cite{koch2019abc} dataset to extract commands from the CSG representations of CAD models via the Onshape API~\cite{Onshape} and FeatureScript, and convert them into parameterized sequences. During the filtering process, we retain only models with complete design operations and discard those with only sketching and extrusion operations, ensuring that the final command sequences have lengths ranging from 60 to 256. We conducted a statistical analysis of the CAD command sequence lengths (see Figure~\ref{fig:datatongji}), which details the distribution of command sequence lengths in the training data. Notably, our dataset, with sequence lengths of 60--256, is currently the longest available CAD command sequence dataset.

\section{Parameterization Details}
\begin{figure}[t]
    \centering
    \includegraphics[width=\linewidth]{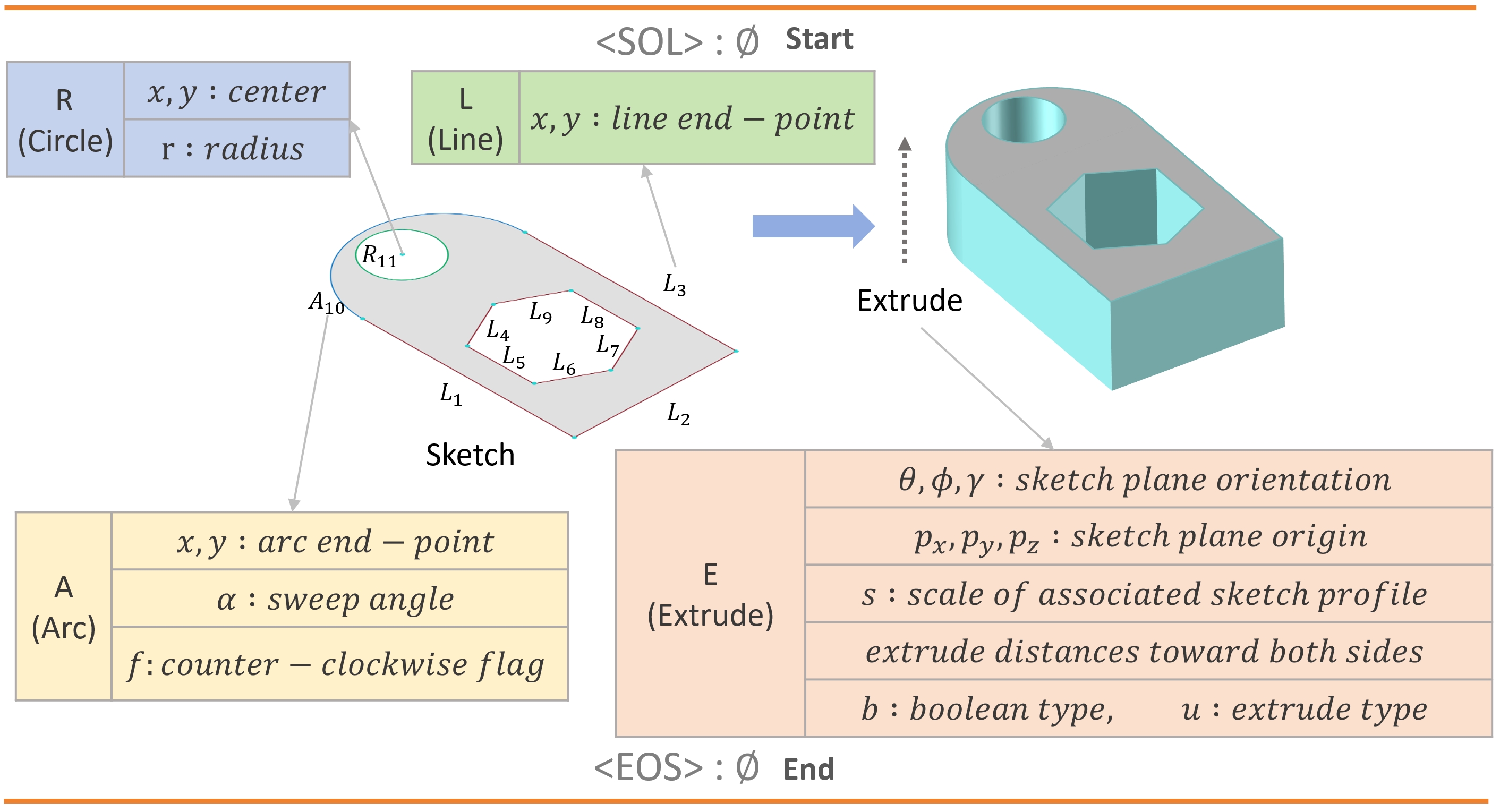}
    \caption{CAD commands and their parameters. $\langle \text{SOL} \rangle$ indicates the start of a loop, $\langle \text{EOS} \rangle$ indicates the end of the entire sequence.}
    \label{fig:command}
\end{figure}

Figure~\ref{fig:command} illustrates the complete command parameter list 
$p_i = [x, y, \alpha, f, r, \theta, \phi, \gamma, p_x, p_y, p_z, s, e_1, e_2, b, u] \in \mathbb{R}^{16}$, which describes the geometric information of each command in a CAD model.
To ensure bounded values, each CAD model is first scaled into a $2\times2\times2$ cube (without translation), which confines the sketch plane origin $(p_x, p_y, p_z)$ and extrusion distances $(e_1, e_2)$ to $[-1,1]$, the sketch profile scale $s$ to $[0,2]$, and the plane directions $(\theta, \phi, \gamma)$ to $[-\pi,\pi]$. Moreover, sketch profiles are normalized to a unit square with the starting point fixed at $(0.5,0.5)$, ensuring that the curve endpoint $(x,y)$ and circle radius $r$ lie in $[0,1]$, and the arc sweep angle $\alpha$ in $[0,2\pi]$. All continuous parameters are discretized into 256 levels using 8-bit integers, while discrete parameters remain unchanged. In a CAD model, sketch commands (initiated by $\langle \text{SOL} \rangle$) and extrusion commands alternate; the former define 2D closed profiles, and the latter extrude these profiles to form 3D solids with specified Boolean operations. Each command is represented as a $16\times1$ vector (unused entries set to $-1$), and the sequence length is fixed at $N_c=256$ (padded with $\langle \text{EOS} \rangle$). This parameterization is similar to that in DeepCAD\cite{wu2021deepcad}.

\section{Model Architecture and Training Details}
\textbf{Autoencoder.} The encoder is composed of four Mamba+ blocks. The state-space model has a feature dimension of 256, a convolution kernel size of 4, and an attention head dimension of 32. The decoder consists of four Transformer blocks, each containing 8 attention heads and a feed-forward network with a dimension of 1024. All modules employ standard layer normalization and a dropout rate of 0.1. The encoder compresses the input sequence into a latent space of dimension 64. At the final stage, the decoder uses two independent linear mappings to predict the command type and the command parameters. The command type prediction produces an output with dimensions 256 by 6 while the command parameter prediction yields an output with dimensions 256 by 4096, which is reshaped into a matrix of size 16 by 256. Training uses the AdamW optimizer with a weight decay set to 1e-4, an initial learning rate of 1e-3, and 2000 warmup steps. The batch size is 32 and the model is trained for 300 epochs with gradient clipping at a threshold of 1.0.

\noindent\textbf{MST-Diffusion Generator.} The diffusion generator adopts a linear diffusion strategy with 1000 steps. The model is trained for 200000 iterations with a batch size of 64. At each step the model predicts noise using mean squared error loss and is optimized by the Adam optimizer with a beta one value of 0.9. The initial learning rate is set to 2e-4 and the learning rate decays after 100000 iterations with a decay factor of 0.1. The diffusion model configuration is consistent with the autoencoder, with a latent vector dimension of 64, a sequence length of 256, an embedding dimension of 512, 6 Transformer layers, 8 attention heads, and a dropout rate of 0.1.

\section{The Algorithm of Mamba+}
As shown in Figure~\ref{fig:Autoencoder_architecture}, the autoencoder architecture incorporates Mamba+ blocks, which utilizes a dual-branch structure to balance local feature extraction and long-range dependency modeling. The input embedding sequence \(E_x \in \mathbb{R}^{B \times W \times D}\) is first decomposed by a linear projection into two branches: a feature transformation branch (b1) and a gating branch (b2). This process produces the main-path feature \(x\) and the gating signal \(z\), respectively. In branch b1, a one-dimensional convolution followed by a SiLU activation enhances local features, yielding \(x'\). Based on \(x'\), the state-space module parameters \(B\) and \(C\), as well as the discretization factor \(\Delta\), are computed. In conjunction with a learnable parameter \(A\), the discretization process generates \(\bar{A}\) and \(\bar{B}\), which are then fed into the state-space model (SSM) to compute \(h_{\mathrm{SSM}}\). Meanwhile, branch b2 applies the Sigmoid function to generate the gating signal \(G_{b2}\); its complement, \(G_{f} = 1 - G_{b2}\), acts as a forget gate over the historical feature \(x'\) to produce the adjusted feature \(x''\). Finally, the output is obtained by fusing \(x''\) with \(h_{\mathrm{SSM}}\) and applying a linear projection to produce \(E_y\), thereby completing the multi-level encoding of the input command sequence. The pseudocode for the Mamba+ algorithm is presented in the following table:

\begin{table}[htbp]
\centering
\resizebox{\columnwidth}{!}{
\begin{tabular}{l}
\toprule
\textbf{Algorithm: Mamba+ Block Implementation} \\
\midrule
\textbf{Input:} Embedded command sequence \( E_x : (B, W, D) \) \\
\textbf{Output:} Encoded features \( E_y : (B, W, D) \) \\
\\[-6pt]
1:\quad \( x \leftarrow \mathrm{Linear}^x(E_x) \) \quad // Feature branch (\( b1 \)) \\
2:\quad \( z \leftarrow \mathrm{Linear}^z(E_x) \) \quad // Gating branch (\( b2 \)) \\
3:\quad \( x' \leftarrow \mathrm{SiLU}(\mathrm{Conv1D}(x)) \) \quad // Local enhancement \\
4:\quad \( A \leftarrow \mathrm{Parameter}^A \quad \scriptstyle{(E \times N)} \) \\
5:\quad \( B \leftarrow \mathrm{Linear}^B(x'), \; C \leftarrow \mathrm{Linear}^C(x') \) \\
6:\quad \( \Delta \leftarrow \log(1 + \exp(\mathrm{Linear}^\Delta(x'))) + \mathrm{Parameter}^\Delta \) \\
7:\quad \( \bar{A}, \bar{B} \leftarrow \mathrm{discretize}(\Delta, A, B) \) \quad // Discretization \\
8:\quad \( h_{\text{SSM}} \leftarrow \mathrm{SSM}(\bar{A}, \bar{B}, C)(x') \) \quad // State-space computation \\
9:\quad \( \mathrm{G}_{b2} \leftarrow \sigma(z), \; \mathrm{G}_f \leftarrow 1 - \mathrm{G}_{b2} \) \quad // Forget gate \\
10:\quad \( x'' \leftarrow \mathrm{G}_f \cdot x' \) \quad // Historical feature retention \\
11:\quad \( h_{\text{out}} \leftarrow x'' + h_{\text{SSM}} \) \quad // Feature integration \\
12:\quad \( E_y \leftarrow \mathrm{Linear}^{y'}(h_{\text{out}}) \) \\
13:\quad \textbf{return} \( E_y \) \\
\bottomrule
\end{tabular}
}
\label{tab:mamba_plus}
\end{table}

\section{Detailed Diffusion Model Architecture}
In this work, we employ a diffusion model guided by a multi-scale Transformer to predict the noise at each diffusion timestep. The model accepts a noisy latent variable 
\(\displaystyle Z_{t}\in\mathbb{R}^{N\times T\times D}\)
and outputs an estimate of the injected noise 
\(\displaystyle \epsilon\). 
A time-conditioning mechanism maps each discrete diffusion step 
\(\displaystyle t\in\{1,\ldots,T\}\)
to a continuous embedding 
\(\displaystyle \tau(t)\), 
which is then processed by an MLP to produce two sets of scale and shift parameters 
\(\{\xi_{1}^l, \psi_{1}^l, \omega_{1}^l,\,\xi_{2}^l, \psi_{2}^l, \omega_{2}^l\}_{l=1}^L\),
where \(L\) is the number of Transformer layers. In particular, \(\xi_i^l\) and \(\psi_i^l\) modulate the layer-normalized activations, whereas \(\omega_i^l\) adjusts the strength of the residual connections.

\noindent\textbf{Multi-Scale Transformer.}
Within each Transformer layer, we introduce three parallel attention branches to capture local, mid-range, and global dependencies. For a position \(i\) in the sequence, we define:
\begin{equation}
\begin{aligned}
W_{l}(i) &= \{\,j \mid |i-j|\le 64 \},\\
W_{m}(i) &= \{\,j \mid |i-j|\le 128 \},\\
W_{g}(i) &= \{\,1,2,\ldots,T\}.
\end{aligned}
\end{equation}
\noindent
We then construct a mask 
\(\displaystyle M_{w}\in\mathbb{R}^{T\times T}\) 
(for \(w\in\{l,m,g\}\)) such that 
\(M_{w}(i,j)=0\)
if 
\(j\in W_{w}(i)\)
and 
\(-\infty\)
otherwise. Each attention branch operates on its respective window, yielding outputs 
\(\displaystyle H_{l},\,H_{m},\,H_{g}\). 
We then fuse these via a learnable gating mechanism:

\begin{equation}
H_{\mathrm{fuse}}
=
W_{o}\Bigl[
\sigma\bigl(W_{g}[\,H_{l}\,\|\,H_{m}\,\|\,H_{g}]\bigr)
\,\odot\,
\bigl(H_{l}\,\|\,H_{m}\,\|\,H_{g}\bigr)
\Bigr],
\end{equation}

\noindent
where \(\|\) denotes concatenation, \(\sigma\) is the sigmoid function, \(\odot\) is elementwise multiplication, and \(W_{g},W_{o}\) are learnable parameters.

\vspace{1ex}
\noindent\textbf{Sequence-Aware Positional Encoding.}
To preserve the sequential logic of CAD commands, we augment each token embedding 
\(\displaystyle Z\in\mathbb{R}^{T\times d}\)
with a learnable scalar \(\eta\). We define:

\begin{equation}
\mathrm{PE}(Z)_{pos,j} 
= 
Z_{pos,j} 
\;+\;
\eta
\,\times\,
\begin{cases}
\sin\Bigl(\tfrac{\mathrm{pos}}{10000^{\,j/d}}\Bigr),
&\text{if }j\text{ is even},\\[1ex]
\cos\Bigl(\tfrac{\mathrm{pos}}{10000^{\,(j-1)/d}}\Bigr),
&\text{if }j\text{ is odd}.
\end{cases}
\end{equation}

\noindent
Let \(\phi_{l}\) denote the \(l\)-th Transformer layer (incorporating multi-scale attention and a feed-forward network). The model estimates the noise via:

\begin{equation}
\epsilon_{\theta}
=
W_{\mathrm{out}}
\circ
\phi_{L}
\circ
\cdots
\circ
\phi_{1}
\Bigl(
\mathrm{PE}\bigl(W_{\mathrm{in}}\,Z_{t}\bigr)
+
\tau(t)
\Bigr).
\end{equation}

\noindent
Inside each layer, we apply layer normalization and inject the time-derived parameters 
\(\{\xi_{i}^l, \psi_{i}^l, \omega_{i}^l\}\)
to modulate the computations. Denoting the layer input by \(X\):

\begin{equation}
\begin{aligned}
X' 
&= 
\mathrm{LayerNorm}(X)\,\odot\,(1+\xi_{1}^l)
\;+\;
\psi_{1}^l,\\[1ex]
X_{\mathrm{attn}}
&=
X
+
\omega_{1}^l\,\odot\,\mathrm{Attention}(X'),\\[1ex]
X'' 
&=
\mathrm{LayerNorm}(X_{\mathrm{attn}})\,\odot\,(1+\xi_{2}^l)
\;+\;
\psi_{2}^l,\\[1ex]
X_{\mathrm{out}}
&=
X_{\mathrm{attn}}
+
\omega_{2}^l\,\odot\,\mathrm{FFN}(X'').
\end{aligned}
\end{equation}

\vspace{1ex}
\noindent\textbf{Forward Diffusion.}
We apply a linear variance schedule \(\{\beta_t\}\):
\begin{equation}
\begin{aligned}
\beta_{t} &= 0.0001 \;+\;0.0199\,\frac{t}{T},\\
Z_{t} &= \sqrt{\alpha_{t}}\;Z_{0}\;+\;\sqrt{1-\alpha_{t}}\;\epsilon,\\
\alpha_{t} &= \prod_{s=1}^{t}\bigl(1-\beta_{s}\bigr),
\end{aligned}
\end{equation}
where \(\epsilon\sim \mathcal{N}(0,I)\). The network is trained to predict \(\epsilon\) by minimizing:
\begin{equation}
\mathcal{L}
=
\mathbb{E}_{t,\,Z_{0},\,\epsilon}
\Bigl[
\|\epsilon - \epsilon_{\theta}(Z_{t},\,t)\|^{2}
\Bigr].
\end{equation}

\vspace{1ex}
\noindent\textbf{Sampling Process.}
Starting from 
\(Z_{T}\sim \mathcal{N}(0,I)\),
we iteratively denoise:
\begin{equation}
Z_{t-1}
=
\frac{1}{\alpha_{t}}
\Bigl(
Z_{t}
-
\beta_{t}
\,\frac{1}{\sqrt{1-\alpha_{t}}}
\,\epsilon_{\theta}(Z_{t},\,t)
\Bigr)
+
\beta_{t}\;z,
\end{equation}
where 
\(z\sim \mathcal{N}(0,I)\).
Here, \(\alpha_{t}\) and \(\beta_{t}\) denote the attenuation factor and variance at step \(t\). This iterative process gradually removes noise to recover a clean latent \(Z_{0}\), enabling faithful reconstruction of the CAD command sequences.

\begin{figure}[!t]
    \centering
    \includegraphics[width=\linewidth]{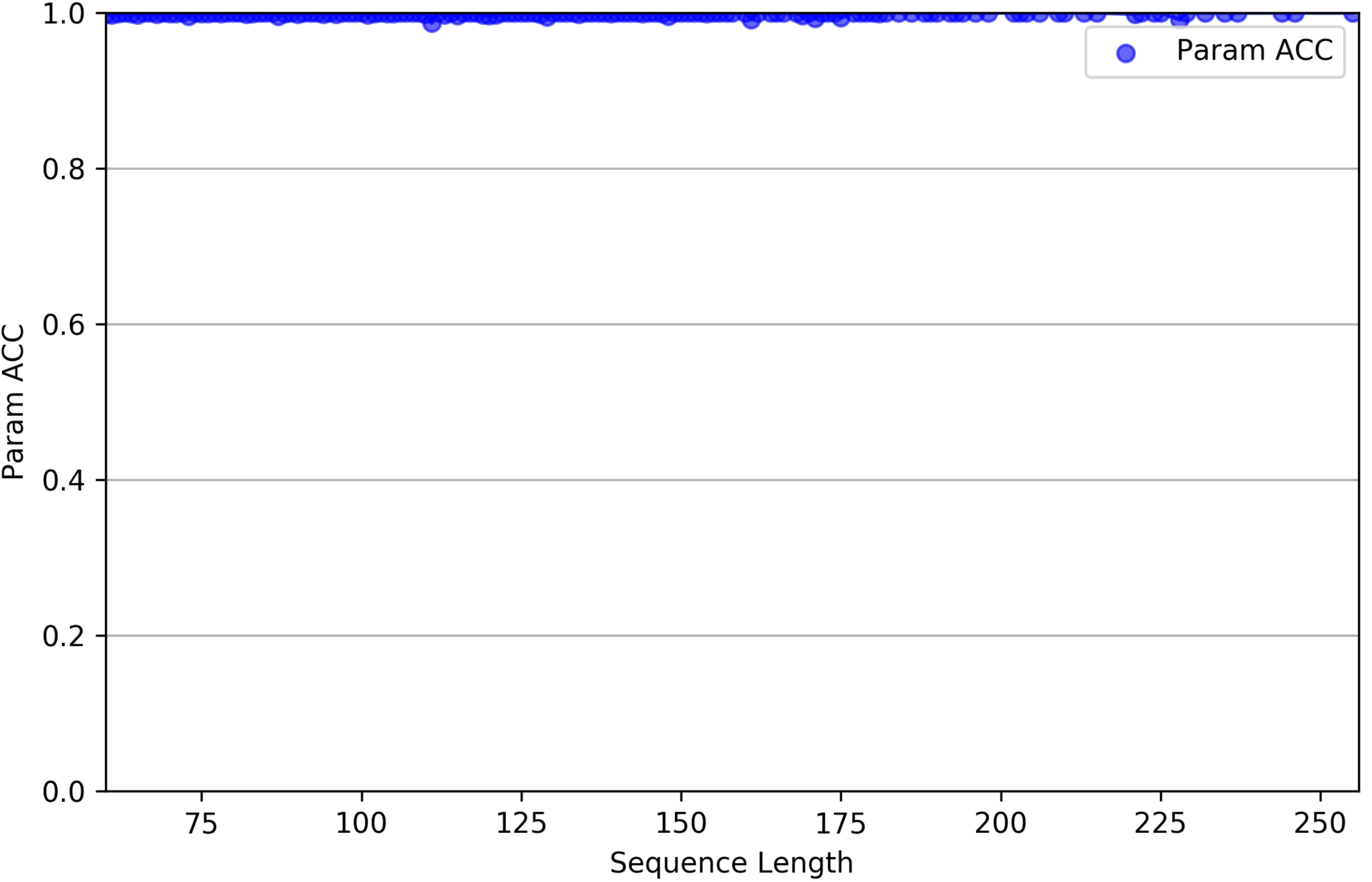}
    \caption{CAD command sequence length and its corresponding parameter accuracy.}
    \label{fig:acc}
\end{figure}

\section{Failure Cases}
\begin{figure}[!t]
    \centering
    \includegraphics[width=\linewidth]{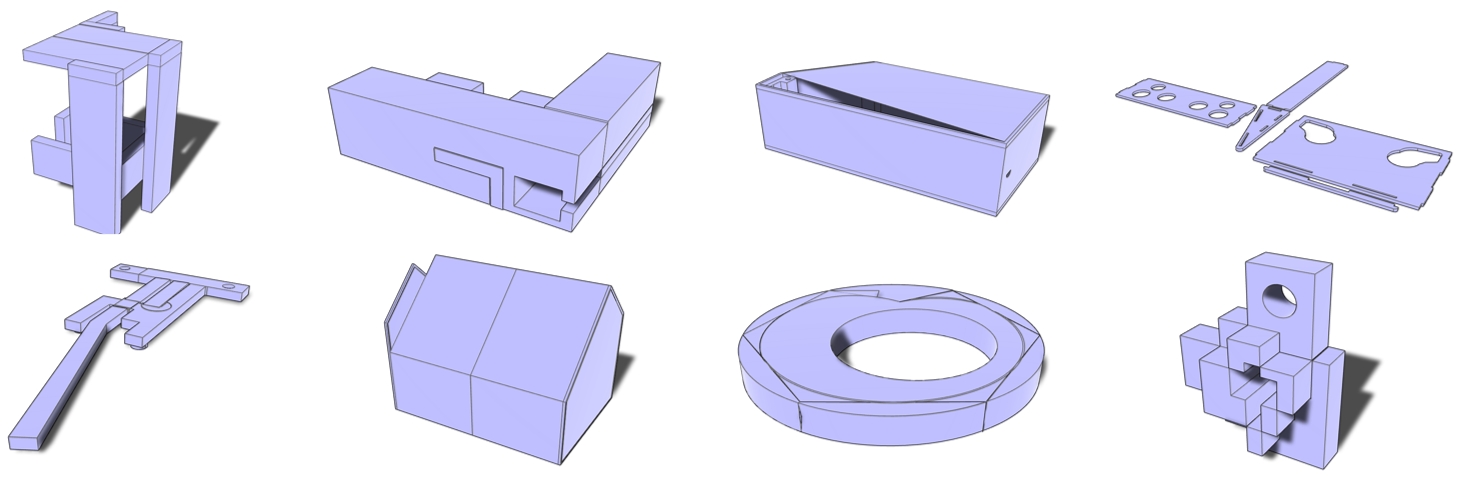}
    \caption{A collection of generated failed CAD models.}
    \label{fig:fail}
\end{figure}

Although our model performs very well on the reconstruction task as shown in Figure~\ref{fig:acc}, maintaining high accuracy on both short and long sequences and achieving parameter prediction accuracy consistently near 0.99, non-watertight objects are occasionally generated during the generation process. Figure~\ref{fig:fail} presents several failure examples, and these failures are mainly caused by limitations in the generator's process.

\begin{figure*}[!t]

    \centering
    \includegraphics[width=1\textwidth]{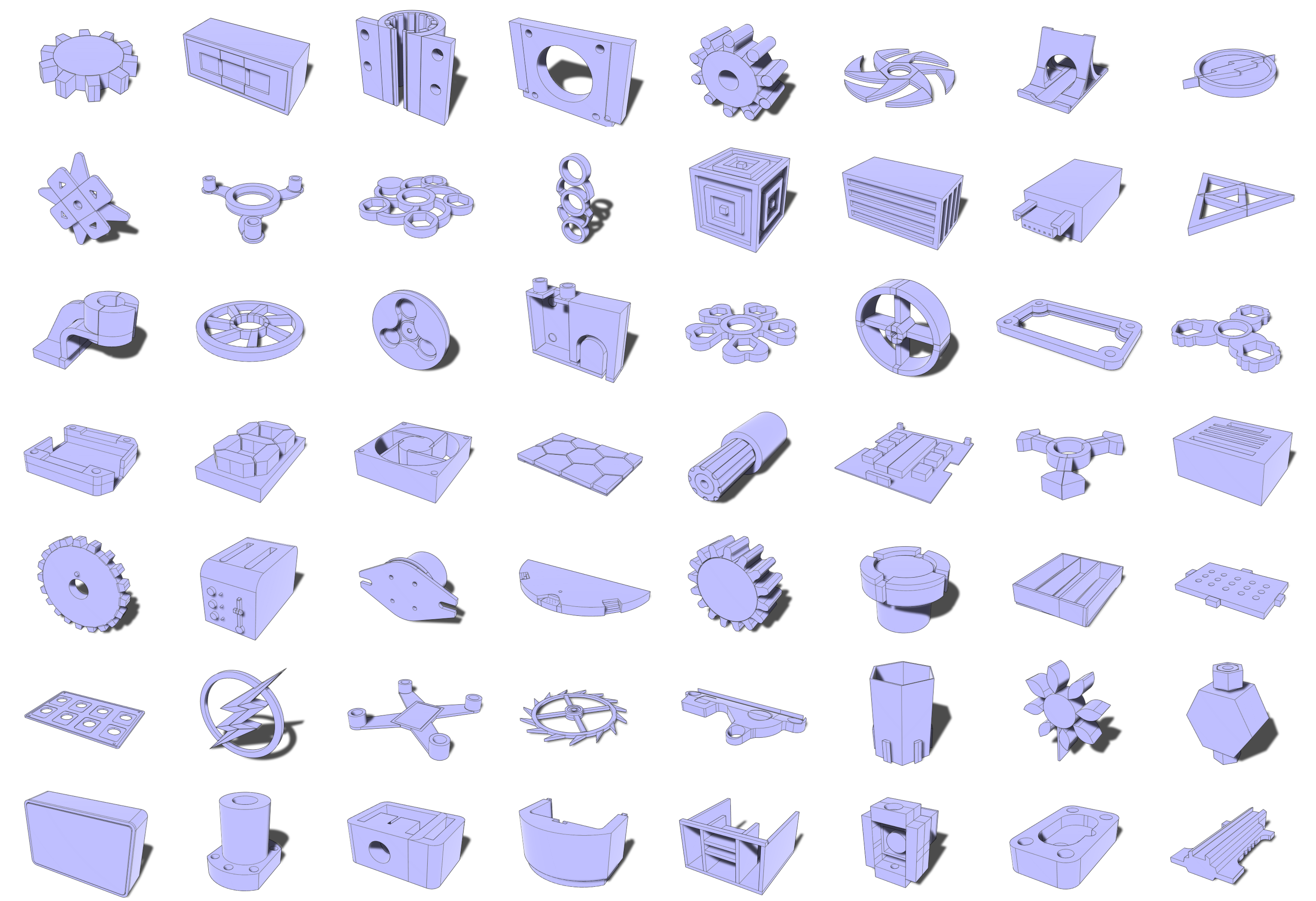}
    \caption{Unconditionally Generated CAD Models}
    \label{fig:quanpingzhanshi}
\end{figure*}

\end{document}